\documentclass[sigconf,nonacm]{acmart}

\usepackage{multirow}
\usepackage{rotating}
\usepackage{graphicx}

\def \hfillx {\hspace*{-\textwidth} \hfill}

\def\blue#1{{\color{blue!80!gray}#1}}
\def\green#1{{\color{green!30!gray}#1}}

\AtBeginDocument{%
  \providecommand\BibTeX{{%
    \normalfont B\kern-0.5em{\scshape i\kern-0.25em b}\kern-0.8em\TeX}}}

\begin{document}

\title{Data Bootstrapping Approaches to Improve Low Resource Abusive Language Detection for Indic Languages}

\author{Mithun Das}
\email{mithundas@iitkgp.ac.in}
\orcid{0000-0003-1442-312X}
\affiliation{%
  \institution{Indian Institute of Technology Kharagpur}
  \state{West Bengal}
  \country{India}
  \postcode{721302}
}

\author{Somnath Banerjee}
\email{som.iitkgpcse@kgpian.iitkgp.ac.in}
\affiliation{%
  \institution{Indian Institute of Technology Kharagpur}
  \state{West Bengal}
  \country{India}
  \postcode{721302}
}

\author{Animesh Mukherjee}
\email{animeshm@cse.iitkgp.ac.in}
\affiliation{%
  \institution{Indian Institute of Technology Kharagpur}
  \state{West Bengal}
  \country{India}
  \postcode{721302}
}

\renewcommand{\shortauthors}{Mithun Das et al.}

\begin{abstract}
Abusive language is a growing concern in many social media platforms. Repeated exposure to abusive speech has created physiological effects on the target users. Thus, the problem of abusive language should be addressed in all forms for online peace and safety. While extensive research exists in abusive speech detection, most studies focus on English. Recently, many smearing incidents have occurred in India, which provoked diverse forms of abusive speech in online space in various languages based on the geographic location. Therefore it is essential to deal with such malicious content. In this paper, to bridge the gap, we demonstrate a large-scale analysis of multilingual abusive speech in Indic languages. We examine different interlingual transfer mechanisms and observe the performance of various multilingual models for abusive speech detection for eight different Indic languages. We also experiment to show how robust these models are on adversarial attacks. Finally, we conduct an in-depth error analysis by looking into the models' misclassified posts across various settings. We have made our code and models public for other researchers\footnote{\url{https://github.com/hate-alert/IndicAbusive}}.
\end{abstract}



\maketitle

\section{Introduction}
Social media platforms (such as Twitter, Facebook, etc.) have connected billions of people at different levels and allowed them to share ideas among themselves instantly. On the one hand, it has led to the exchange of thoughts and massive expansion of social networks; on the other hand, these platforms have been used to spread propaganda, violence, and abuse against users based on gender, religion, race, geographic location, etc.~\cite{das2020hate}. Not only such abusive behavior can lead to the traumatization of the victims by affecting them psychologically~\cite{vedeler2019hate}, but these can also ignite social tensions and affect the stature of the platforms which host them~\cite{statt2017youtube}. Further, widespread usage of such content can also have implications in the offline world: violent hate crimes,  youth suicides, mass shootings, and extremist recruitment~\cite{johnson2019hidden}.

To mitigate the spread of such abusive content, these platforms have come up with specific guidelines\footnote{\label{twitter_violation}\url{https://help.twitter.com/en/rules-and-policies/hateful-conduct-policy}} that need to be followed by the users of these platforms. Failure to follow these guidelines may result in the deletion of the post or suspension of the user's account. To reduce the hateful content from these platforms, they employ moderators~\cite{newton_2019} to review posts manually and keep the forum healthy and friendly. However, this moderation technique is limited by the moderators' speed, diction, ability to understand the evolution of vernacular, and familiarity with multilingual content. In addition, many moderators complain about psychological effects induced due to moderation of such abusive content\cite{the_guardian_2017}. Besides, due to the sheer volume and velocity of data streaming, it is an ambitious attempt to filter all the posts manually and screen out such hostile content. Thus, automatic detection of such abusive content is incredibly crucial and unavoidable.

It has already been eyed that Facebook actively removed a large portion of malicious content from their platform even before users flagged it\footnote{\url{https://time.com/5739688/facebook-hate-speech-languages/}}. Nevertheless, the limitation is that these platforms can identify such detrimental content in certain major languages such as English, Spanish, etc. To that end, several studies have been performed for automated detection of abusive content, concentrating predominantly on the English language. Hence, an effort is needed to determine and mitigate such malicious content in low-resource languages.

In the last few years, there have been a series of incidents in India, such as smearing movements against famous political leaders\footnote{\url{https://nenow.in/top-news/resign-pm-modi-trends-on-twitter-as-india-sees-worst-covid-19-wave.html}}, celebrities\footnote{\url{https://www.koimoi.com/bollywood-news/sushant-singh-rajput-news-rhea-chakraborty-was-harassed-by-various-agencies-satish-maneshinde-demands-cbi-findings}}, and social media personalities, online anti-religious propaganda\footnote{\url{https://www.news18.com/news/buzz/indians-wants-to-boycott-myntra-for-old-anti-hindu-poster-it-didnt-even-make-4115855.html}}, cyber harassment\footnote{\url{https://timesofindia.indiatimes.com/city/mumbai/cyber-harassment-cases-see-upswing-in-pandemic/articleshow/88842765.cms}}, etc. So, to deal with such malicious content, automated systems are much required to keep the online ecosystem healthy. India has more than 1.3 billion people, having the highest number of users on Facebook\footnote{\url{https://worldpopulationreview.com/country-rankings/facebook-users-by-country}}, YouTube\footnote{\url{https://www.globalmediainsight.com/blog/youtube-users-statistics/}} and the third-highest number of users on Twitter\footnote{\url{https://www.statista.com/statistics/242606/number-of-active-twitter-users-in-selected-countries/}}. Besides, the country has 22 recognized languages, which are spoken in various parts of it\footnote{https://www.universal-translation-services.com/recognized-languages-in-india/}. Due to the most extensive language diversity and their usage on social media, detecting such abusive content in all languages becomes challenging. There are primarily two reasons for this -- annotators need to have diverse language expertise and should have knowledge in abusive content analysis. 

Further, a recent trend on social media platforms is that people write non-English languages using English characters and switch among two or more languages in the same conversation. This phenomenon is called code-mixing (or code-switching), where linguistic units such as phrases, words, or morphemes of one language are embedded in an utterance of another language. Code-mixing allows ease of communication among speakers by furnishing a more comprehensive set of phrases and expressions. However, this has also made the task of developing NLP tools more complex, as emphasized by Chittaranjan et al.~\cite{chittaranjan2014word}. Therefore, there is a need to develop efficient models to detect abusive content in Indic languages in various settings.

This paper performs a large-scale analysis of multilingual abusive speech by investigating multilingual models' performance in eight different Indic languages. We address the question of data scarcity and language similarity/typology, two under-explored issues in abusive language identification. Inspired by Nag et al.\cite{nag2021data}, we explore and catalog a variety of strategies for transferring abusive language detection ability across languages, especially in the resource-rich to resource-poor direction, starting from the ``each language for itself" (ELFI) criteria. To investigate the degree to which various transfer modes can compensate for gold training instances, we explore multiple scenarios ranging from zero-shot learning, few-shot learning, instance transfer, cross-lingual learning, etc. (more details in section \ref{sec:method}). In summary, we observe that --
\begin{itemize}
    \item  In ELFI style training, although \textbf{m-BERT} is pre-trained in more than 100 languages, \textbf{MuRIL} outperforms m-BERT in 7 out of 10 language types as it is pre-trained explicitly in Indian languages.
    \item The zero-shot model transfer can be beneficial for abusive language detection when the source and the target languages belong to the same language family. For few-shot settings, \textbf{\textit{AllBOne}} is the most effective.
    \item The model transfer brings better performance than instance transfer (i.e., all instances of a resource-poor language are first translated to a resource-rich language and the training and predictions are done in the resource-rich language) due to the added translation error in instance transfer which reduces the overall toxicity score of an abusive post.
    \item For low-resource languages, though synthetic silver instances are helpful in detecting abusive content up to some degree, further fine-tuning with gold target instances gives steady improvements.
    \item Our in-depth error analysis reveals that when a post's contextual information is limited, implicit, or discriminatory features are present, the model fails to detect such an abusive post.
\end{itemize}

\section{Related Works}

\subsection{Dynamics of online abuse}

Online hostility is a context-dependent notion intended to express hatred and threaten an individual or group based on discriminatory views. Despite the argument that hateful statements ought to be tolerated due to free speech acts, the public expression of hate speech propels the reduction of minority members, and such frequent and repetitive exposure to abusive speech could increase an individual's outgroup prejudice~\cite{soral2018exposure}. Real-world violent events could also lead to increased hatred in online space and vice versa~\cite{olteanu2018effect}. With the rise of online hate, the research community has a massive responsibility to develop solutions to mitigate online hostility.

\subsection{Research on abusive speech}
The concern of abusive speech has long been studied in the research community. Earlier work on abusive speech attempted to detect abusive users by using lexical, syntactic features extracted from their posts~\cite{chen2012detecting}. Over the past few years, research around automated hate speech detection has matured tremendously. Most of the current study consists of diverse but related works. In 2016, Zeerak Waseem and Dirk Hovy~\shortcite{waseem2016hateful} contributed a dataset in which thousands of tweets were labeled with racism and sexism markers, and Davidson et al.~\shortcite{davidson2017automated} focused on distinguishing offensive from hate content on Twitter. Using this dataset, the authors examined multiple linguistic features such as character and word n-grams, POS tags, emotion lexicon, and tf-idf vectors with several classifiers such as LR, SVM, decision tree, etc. Although multiple datasets were being published, a major problem was the lack of correlation and re-usable datasets across hate speech detection tasks~\cite{kumar2018benchmarking}. To address this issue, Founta et al.~\shortcite{founta2018large} studied these inconsistencies and drew upon a robust labeling mechanism that attempts to circumvent the overlap among various forms of abusive speech.

With the advent of large datasets, most academic research has moved to data-hungry complex models to improve classifier performance, including deep learning~\cite{badjatiya2017deep} and graph embedding techniques~\cite{das2021you}.  Pitsilis et al.~\cite{Pitsilis2018DetectingOL}, used deep learning models such as LSTMs to identify the abusive tweets in English and noticed that it was pretty effective in this task. Zhang et al.~\cite{zhang2018detecting} fused convolutional and gated recurrent networks to enhance the classification performance and had remarkable success on 6 out of 7 datasets used. Recently, transformer based language models such as BERT are becoming immensely popular in several downstream tasks and have outperformed several deep learning models such as CNN-GRU, LSTM, etc., for detecting abusive language~\cite{banerjee2021exploring,das2021abusive}.

\subsection{Abusive language detection in Indic languages}

In the last few years, several shared tasks, such as Hate-Speech and Offensive Content Identification (HASOC)~\cite{mandl2021overview}, Dravidian Lang-Tech~\cite{chakravarthi2021findings} workshop, TRAC~\cite{kumar2018proceedings}, etc., have been organized to develop resources, datasets, and models for abusive speech detection and multiple datasets in Indic languages such as Hindi, Marathi, Tamil, Malayalam, etc. have been made public. The HASOC~\cite{mandl2021overview} shared task in Indo-European languages is arguably the most well-known series of competitions. It has been consistently organized from 2019 at the Forum for Information Retrieval (FIRE). The Dravidian Lang-Tech~\cite{chakravarthi2021findings} workshop focused on determining the offensive language of the code-mixed dataset in three Dravidian languages, namely, Tamil–English, Malayalam–English, and Kannada–English crawled from social media. In addition, researchers have also developed several datasets for Bengali~\cite{romim2021hate}, code-mixed Hindi~\cite{bohra2018dataset}, Urdu~\cite{akhter2020automatic,rizwan2020hate}, etc., for abusive language detection. However, a limited number of studies have been performed on the effect of zero-shot learning, few-shot learning~\cite{ranasinghe2020multilingual,ranasinghe2021evaluation}, instance transfer, etc. In our work, we try to fill this critical gap by studying various transfer schemes thus opening up new avenues for future research for abuse detection in Indic languages.

\begin{table*}
\begin{tabular}{|l|l|l|l|l|l|l|}
\hline
\textbf{Language} & \textbf{Type} & \textbf{Dataset} & \textbf{Source} & \textbf{Abusive} & \textbf{Normal} & \textbf{Total} \\ \hline
\textbf{Bengali} & \textbf{\textit{Bn}} & \textit{Romin et al.}\cite{romim2021hate} & Facebook \& YouTube & 10,000 & 20,000 & 30,000 \\ \hline
\multirow{3}{*}{\textbf{English}} & \multirow{3}{*}{\textbf{\textit{En}}} & \textit{Davidson et al.}~\cite{davidson2017automated}  & Twitter & 20,620 & 4,163 & 24,783 \\ \cline{3-7} 
 &  & \textit{Founta et al.}~\cite{founta2018large} & Twitter & 31,985 & 53,790 & 85,775 \\ \cline{3-7}  
 &  & \textit{HateXplain}~\cite{mathew2021hatexplain} & Twitter \& Gab & 11,415 & 7,814 & 19,229 \\ \hline
\multirow{4}{*}{\textbf{Hindi}} & \multirow{3}{*}{\textbf{\textit{Hi}}} & \textit{Mandl et al.}~\cite{mandl2019overview} & Twitter \& Facebook & 3,074 & 2,909 & 5,983 \\ \cline{3-7} 
 &  & \textit{Mandl et al.}~\cite{mandl2020overview} & Twitter & 1,044 & 2,582 & 3,626 \\ \cline{3-7} 
 &  & \textit{Mandl et al.}~\cite{mandl2021overview} & Twitter & 1,938 & 4,188 & 6,126 \\ \cline{2-7} 
 & \textbf{\textit{Hi-En}} & \textit{Bohra et al.}~\cite{bohra2018dataset} & Twitter & 1,661 & 2,918 & 4,579 \\ \hline
\textbf{Kannada} & \textbf{\textit{Ka-En}} & \textit{Chakravarthi et al.}~\cite{chakravarthi2021findings} & YouTube & 1,465 & 4,188 & 5,873 \\ \hline
\multirow{2}{*}{\textbf{Malayalam}} & \multirow{2}{*}{\textbf{\textit{Ma-En}}} & \textit{Chakravarthi et al.}~\cite{chakravarthi2021findings} & YouTube & 706 & 17,697 & 18,403 \\ \cline{3-7} 
 &  & Mandl et al.~\cite{mandl2020overview} & YouTube & 2,430 & 2.520 & 4,950 \\ \hline
\textbf{Marathi} & \textbf{\textit{Mr}} & \textit{Gaikwad et al.}~\cite{gaikwad2021cross} & Twitter & 865 & 1,611 & 2,499 \\ \hline
\textbf{Tamil} & \textbf{\textit{Ta-En}} & \textit{Chakravarthi et al.}~\cite{chakravarthi2021findings} & YouTube & 12,651 & 33,684 & 47,072 \\ \hline
\multirow{4}{*}{\textbf{Urdu}} & \multirow{2}{*}{\textbf{\textit{Ur}}} & \textit{Akhter et al.}~\cite{akhter2020automatic}  & YouTube & 1,109 & 1,062 & 2,171 \\ \cline{3-7} 
 &  & \textit{Amjad et al.}~\cite{hasoc2021fireabuse} & Twitter & 1,750 & 1,750 & 3,500 \\ \cline{2-7} 
 & \multirow{2}{*}{\textbf{\textit{Ur-En}}} & \textit{Khan et al.}~\cite{khan2021hate} & Twitter & 3,575 & 1,425 & 5,000 \\ \cline{3-7} 
 &  & \textit{Rizwan et al.}~\cite{rizwan2020hate} & Twitter & 5,349 & 4,664 & 10,013 \\ \hline
\textbf{Total} & \textbf{-} & \textit{-} & - & 111,637 & 159,859 & 271,496 \\ \hline
\end{tabular}
\caption{Details of the datasets.}
\label{tab:datasetDetail}
\end{table*}

\section{Datasets}

In this section, we describe the datasets used in this paper. We looked into several datasets for Indic languages for abusive speech detection and attempted to gather all of them. To this purpose, we accumulated datasets in 8 (10 types) different languages from 14 publicly available sources. These datasets differ in their choice of class labels (\textit{offensive}, \textit{hate-speech}, \textit{normal}). Apart from normal posts, we combine all other labels as abusive to pose the problem as a binary classification task. The details about the datasets are presented in Table \ref{tab:datasetDetail}.

\noindent\textbf{Bengali (\textit{Bn}):} The dataset shared by Romin et al. \cite{romim2021hate} is one of the largest in Bengali, consisting of 30K posts, among which 10K posts are hateful, and 20K posts are normal. The dataset has been developed by extracting comments from YouTube and Facebook pages. The author employed 50 annotators and instructed them with proper guidelines to build the dataset. To validate the annotation quality, the author randomly sampled 300 posts for each annotator, manually reannotated them, and found that 91.05\%  annotation was correct.

\noindent\textbf{English (\textit{En}):} The majority of the abusive speech datasets are available in the English language, and out of these, we select only three publicly available popular datasets. The work on automatic hate speech detection by Davidson et al.~\cite{davidson2017automated} made public a Twitter dataset consisting of 24k tweets. To curate the dataset, they used a set of lexicons derived from Hatebase.org\footnote{\url{https://hatebase.org/}}. Each tweet has been labeled as either hate speech, offensive or normal. Founta et al.~\cite{founta2018large} shared a large-scale Twitter dataset consisting of more than 100K tweets. Each tweet has been labeled into one of the following categories: hateful, abusive, normal, and spam. The tweets were annotated by 5-20 annotators to maintain the quality of the labels. We ignore the data points annotated as spam from our analysis. The HateXplain dataset introduced by Mathew et al.~\cite{mathew2021hatexplain} is a recent addition to the hate speech research community. It contains around 20K posts categorized into three labels: hate speech, offensive, or normal. The dataset is collected from Twitter and Gab.

\noindent\textbf{Hindi:} For the Hindi language, we found two types of datasets: Devanagari Hindi and code-mixed Hindi. The difference between Devanagari and code-mixed is, though the semantic expression of both types of posts is almost similar, Devanagari Hindi is written using the Devanagari script while code-mixed Hindi is written using English characters. Due to differences in writing style, we treat these languages separately.

\begin{itemize}
    \item [-] \textbf{Devanagari Hindi (\textit{Hi}):} In 2019, Mandl et al.~\cite{mandl2019overview} released a dataset of 5.9K tweets via the HASOC shared task. The posts were labeled into one of the following classes: hate, offensive, profane and normal. The author developed the dataset by crawling tweets and comments from Twitter and Facebook. Following the previous work, in 2020, Mandl et al.~\cite{mandl2020overview} shared another Hindi abusive speech dataset of 3.6K. Recently, the author introduced another dataset~\cite{mandl2021overview} of 6.1K due to the huge success of the previously organized shared task. We combine all three datasets to have our final Hindi dataset.
    
    \item [-] \textbf{Code-mixed Hindi (\textit{Hi-En}):} Bohra et al.~\cite{bohra2018dataset} introduced the first code-mixed Hindi hate speech Twitter dataset. Each tweet has been annotated as either hate speech or non-hate speech. According to their annotation guidelines, they considered any kind of abuse as hateful. The final dataset consists of 4.5K tweets, out of which 1.6K tweets are hateful, and the remaining 2.9K are non-hateful.
\end{itemize}

\noindent\textbf{Kannada (\textit{Ka-En}):} Chakravarthi et al.~\cite{chakravarthi2021findings} released an offensive language identification dataset in Kannada-English (i.e., code-mixed Kannada). To develop the dataset, the author crawled YouTube comments in 2019. The comments are labeled into one of the following categories: not-offensive, offensive-untargeted, offensive-targeted-individual, offensive-targeted-group, offensive-targeted-other, and not-in-indented-language. The final dataset consists of around 7.7K comments. We removed the data points labeled as not-in-indented-language for our analysis, which was irrelevant. All the not-offensive points are assumed to be in the normal class while all the other points (except not-in-indented-language) are fused to form the abusive class.

\noindent\textbf{Malayalam (\textit{Ma-En}):} Similar to the Kannada dataset, Chakravarthi et al. ~\cite{chakravarthi2021findings} made public another code-mixed offensive language detection dataset in Malayalam. The dataset consists of around 20K comments, out of which around 700 comments are offensive. Mandl et al.~\cite{mandl2020overview} made public another code-mixed Malayalam dataset consisting 4.9K comments out of which 2.4K comments are abusive.

\noindent\textbf{Marathi (\textit{Mr}):} The Marathi dataset shared by Gaikwad et al.~\cite{gaikwad2021cross} consists of 2.4K posts, among which 1.62K posts are labeled as offensive, and the remaining posts are marked as non-offensive. The dataset has been curated by extracting tweets from Twitter by searching common curse words in Marathi and phrases related to politics, entertainment, and sports. The author employed six volunteer annotators who were native speakers of Marathi with ages between 20 and 25 years old and a bachelor's degree.

\noindent\textbf{Tamil (\textit{Ta-En}):} Chakravarthi et al. ~\cite{chakravarthi2021findings} made public another offensive language detection dataset in code-mixed Tamil. The dataset consists of 43K comments, making it one of the most extensive datasets in code-mixed Tamil.

\noindent\textbf{Urdu:} We also found two types (actual and code-mixed) of datasets for the Urdu language.

\begin{itemize}
    \item [-] \textbf{Actual Urdu (\textit{Ur}):}  For the Urdu language, we found two publicly available datasets. Akhter et al.~\cite{akhter2020automatic} shared a dataset consisting of 2.1K (1.1K offensive, 1K non-offensive) comments scrapped from YouTube videos. The dataset was manually annotated by three annotators who are native speakers of Urdu. Amjad et al.~\cite{hasoc2021fireabuse} made public another Urdu abusive dataset crawled from Twitter, consisting of 3.5K tweets, out of which 1.75K tweets are abusive, and the remaining are labeled as non-abusive.
    
    \item [-] \textbf{Code-mixed Urdu (\textit{Ur-En}):} For code-mixed Urdu, we use the following two datasets. Khan et al.~\cite{khan2021hate} released a dataset of 5K tweets. Each tweet has been annotated into one of the following categories: hate, offensive, neutral. Rizwan et al.~\cite{rizwan2020hate} made public another code-mixed Urdu dataset having 10K posts containing five different classes: offensive, sexism, religious hate, profane, and normal. Excluding normal, we map all other classes into one label, i.e., abusive.
\end{itemize}

\section{Methodology}
\label{sec:method}

\subsection{Base models}

\noindent\textbf{m-BERT(MB)}: m-BERT~\cite{Devlin2019BERTPO} is pre-trained on 104 languages with the largest Wikipedia utilizing a masked language modeling (MLM) objective. It is a stack of transformer encoder layers with 12 ``attention heads," i.e., fully connected neural networks augmented with a self-attention mechanism. m-BERT is restricted in the number of tokens it can handle (512 at max). To fine-tune m-BERT, we also add a fully connected layer with the output corresponding to the CLS token in the input. This CLS token output usually holds the representation of the sentence passed to the model. The m-BERT model has been well studied in abusive speech, has already surpassed existing baselines, and stands as a state-of-the-art.

\noindent\textbf{MuRIL(MU):} MuRIL~\cite{khanuja2021MuRIL} stands for Multilingual Representations for Indian Languages and aims to improve interoperability from one language to another. This model uses a BERT base architecture pretrained from scratch utilizing the Wikipedia, Common Crawl, PMINDIA, and Dakshina corpora for 17 Indian languages and their transliterated counterparts.

\begin{table*}[t]
        \begin{minipage}{0.4\textwidth}
            \centering
\begin{tabular}{|c|c|c|c|}
\hline
\textbf{Lang}                   & \textbf{Model} & \textbf{Accuracy}   & \textbf{M-F1}  \\ \hline
\multirow{2}{*}{\textbf{\textit{Bn}}}    & \textbf{MB}  & 0.903          & 0.892          \\ \cline{2-4} 
                                & \textbf{MU}  & \textbf{0.905} & \textbf{0.894} \\ \hline
\multirow{2}{*}{\textbf{\textit{Hi}}}    & \textbf{MB}  & 0.776          & 0.768          \\ \cline{2-4} 
                                & \textbf{MU}  & \textbf{0.806} & \textbf{0.799} \\ \hline
\multirow{2}{*}{\textbf{\textit{Hi-En}}} & \textbf{MB}  & \textbf{0.724} & 0.673          \\ \cline{2-4} 
                                & \textbf{MU}  & 0.722          & \textbf{0.693} \\ \hline
\multirow{2}{*}{\textbf{\textit{Ka-En}}} & \textbf{MB}  & 0.783          & 0.736          \\ \cline{2-4} 
                                & \textbf{MU}  & \textbf{0.824}          & \textbf{0.778}          \\ \hline
\multirow{2}{*}{\textbf{\textit{Ma-En}}} & \textbf{MB}  & \textbf{0.902} & \textbf{0.815} \\ \cline{2-4} 
                                & \textbf{MU}  & 0.899          & \textbf{0.815} \\ \hline
\multirow{2}{*}{\textbf{\textit{Mr}}}    & \textbf{MB}  & 0.885          & 0.872          \\ \cline{2-4} 
                                & \textbf{MU}  & \textbf{0.895} & \textbf{0.887} \\ \hline
\multirow{2}{*}{\textbf{\textit{Ta-En}}} & \textbf{MB}  & 0.829          & 0.791          \\ \cline{2-4} 
                                & \textbf{MU}  & \textbf{0.839} & \textbf{0.798} \\ \hline
\multirow{2}{*}{\textbf{\textit{Ur-En}}} & \textbf{MB}  & 0.800          & 0.794          \\ \cline{2-4} 
                                & \textbf{MU}  & \textbf{0.816} & \textbf{0.810} \\ \hline
\multirow{2}{*}{\textbf{\textit{Ur}}}    & \textbf{MB}  & \textbf{0.902} & \textbf{0.902} \\ \cline{2-4} 
                                & \textbf{MU}  & 0.895          & 0.895          \\ \hline
\multirow{2}{*}{\textbf{\textit{En}}}    & \textbf{MB}  & 0.917          & \textbf{0.917} \\ \cline{2-4} 
                                & \textbf{MU}  & \textbf{0.918} & \textbf{0.917} \\ \hline
\end{tabular}
\caption{Performance of ELFI using \textbf{MB} and \textbf{MU}. (Best performance is highlighted using bold).}
\label{tab:indRes}
        \end{minipage}
        \hfillx
        \begin{minipage}{0.5\textwidth}
            \centering
            \begin{tabular}{|c|c|cl|cl|}
            \hline
            \textbf{Lang} & \textbf{Model} & \multicolumn{2}{c|}{\textbf{Accuracy}} & \multicolumn{2}{c|}{\textbf{M-F1}} \\ \hline
            \multirow{2}{*}{\textbf{\textit{Bn}}} & \textbf{MB} & \textbf{0.906} & (+0.003) & \textbf{0.896} & (+0.004) \\ \cline{2-6} 
            & \textbf{MU} & \textbf{0.906} & (+0.001) & 0.895 & (+0.001) \\ \hline
            \multirow{2}{*}{\textbf{\textit{Hi}}} & \textbf{MB} & 0.799 & (+0.023) & 0.783 & (+0.015) \\ \cline{2-6} 
            & \textbf{MU} & \textbf{0.804} & (-0.002) & \textbf{0.794} & (-0.005) \\ \hline
            \multirow{2}{*}{\textbf{\textit{Hi-En}}} & \textbf{MB} & \textbf{0.692} & (-0.032) & 0.605 & (-0.068) \\ \cline{2-6} 
            & \textbf{MU} & 0.636 & (-0.086) & \textbf{0.622} & (-0.071) \\ \hline
            \multirow{2}{*}{\textbf{\textit{Ka-En}}} & \textbf{MB} & 0.803 & (+0.02) & 0.720 & (-0.016) \\ \cline{2-6} 
            & \textbf{MU} & \textbf{0.836} & (+0.012) & \textbf{0.771} & (-0.007) \\ \hline
            \multirow{2}{*}{\textbf{\textit{Ma-En}}} & \textbf{MB} & 0.891 & (-0.011) & 0.797 & (-0.018) \\ \cline{2-6} 
            & \textbf{MU} & \textbf{0.894} & (-0.005) & \textbf{0.802} & (-0.013) \\ \hline
            \multirow{2}{*}{\textbf{\textit{Mr}}} & \textbf{MB} & 0.889 & (+0.004) & 0.877 & (+0.005) \\ \cline{2-6} 
            & \textbf{MU} & \textbf{0.909} & (+0.014) & \textbf{0.901} & (+0.014) \\ \hline
            \multirow{2}{*}{\textbf{\textit{Ta-En}}} & \textbf{MB} & 0.836 & (+0.007) & 0.786 & (-0.005) \\ \cline{2-6} 
            & \textbf{MU} & \textbf{0.845} & (+0.006) & \textbf{0.795} & (-0.003) \\ \hline
            \multirow{2}{*}{\textbf{\textit{Ur-En}}} & \textbf{MB} & 0.763 & (-0.037) & 0.754 & (-0.04) \\ \cline{2-6} 
            & \textbf{MU} & \textbf{0.772} & (-0.044) & \textbf{0.763} & (-0.047) \\ \hline
            \multirow{2}{*}{\textbf{\textit{Ur}}} & \textbf{MB} & 0.905 & (+0.003) & 0.905 & (+0.003) \\ \cline{2-6} 
            & \textbf{MU} & \textbf{0.906} & (+0.011) & \textbf{0.906} & (+0.011) \\ \hline
            \multirow{2}{*}{\textbf{\textit{En}}} & \textbf{MB} & \textbf{0.915} & (-0.002) & \textbf{0.915} & (-0.002) \\ \cline{2-6} 
            & \textbf{MU} & 0.912 & (-0.006) & 0.912 & (-0.005) \\ \hline
        \end{tabular}
        \caption{Performance of joint training. Best performance is highlighted using bold and values within parenthesis denote improvements over to ELFI.}
        \label{tab:jointTraing}
    \end{minipage}
\end{table*}

\subsection{Interlingual transfer mechanisms}

One of the primary interests of transformer-based models is their potential to leverage model transfer via several mechanisms. This can be especially helpful for improving the performance of learning in low-resource languages like Bengali, Hindi, Urdu, etc. We perform the following tests to evaluate the extent to which language similarity can boost transfer learning performance.

\subsubsection{ELFI (Each language for itself)}
We use data from the same language for training, validation, and testing in this setting. This scenario usually occurs in the real world, where annotated (labeled) datasets are used to create classifications for a specific language. While the labeling costs are potentially high, this provides an idea of the most feasible classification performance.

\subsubsection{Joint training/Cross-lingual training}
In this technique, we combine the datasets of all the languages for training the transformer-based models. The idea is that even though the characters, words used to represent different languages vary, the contextual representation of these abusive posts is the same to good extent. In specific, we consider pre-trained embeddings of all the datapoints from all languages (inclusive of the target language) and test it on the test data of the target language. Thus, it gives an idea of whether jointly training the models can help learn a particular post's better semantic representation for determining its corresponding label. 

\subsubsection{Model transfer}
In this setting, the models are trained with one language (source language) and assessed on another language (target language). In the zero-shot setting, no instances from the target language have been used while training  (\textbf{MTx0}). In a related few-shot setting, we allow $n$ = 32 and 64 posts per label from the available gold target instances to fine-tune the current models (trained on another language). These are called \textbf{MTx32} and  \textbf{MTx64}. Another extended variant of this model is where for a target language, we use the dataset of all other languages (combined source) to train the models and evaluate their performance on the target language. It would be the case in which we would like to deploy an abusive speech classifier for a target language directly which does not have any training instances. We name the language model as \textbf{\textit{AllBOne}}.

\begin{table*}
\scriptsize
\centering
\begin{tabular}{|c|c|c|c|c|c|c|c|c|c|c|c|c|} 
\hline
\begin{tabular}[c]{@{}c@{}}\\\\\textbf{Lang}\end{tabular} & \textbf{Model} & \textbf{\textit{Bn}} & \textbf{\textit{Hi}} & \textbf{\textit{Hi-En}} & \textbf{\textit{Ka-En}} & \textbf{\textit{Ma-En}} & \textbf{\textit{Mr}} & \textbf{\textit{Ta-En}} & \textbf{\textit{Ur-En}} & \textbf{\textit{Ur}} & \textbf{\textit{En}} & \textbf{\textit{AllBOne}}  \\ 
\hline \hline
\multirow{2}{*}{\textbf{\textit{Bn}}}                              & \textbf{MB}    & -           & \green{0.670}       & 0.431          & \textbf{0.641}          & 0.482          & 0.631       & 0.617          & 0.368          & 0.461       & 0.403       & \textbf{\blue{0.678}}             \\ 
\cline{2-13}
                                                          & \textbf{MU}    & -           & \textbf{\green{0.713}}       & \textbf{0.497}          & 0.635          & \textbf{0.633}          & \textbf{0.698}       & \textbf{0.625}          & \textbf{0.415}          & \textbf{\blue{\underline{0.763}}}       & \textbf{0.510}       & 0.652             \\ 
\hline \hline
\multirow{2}{*}{\textbf{\textit{Hi}}}                              & \textbf{MB}    & 0.391       & -           & \textbf{0.443}          & \blue{0.573}          & 0.434          & 0.514       & 0.499          & \textbf{0.335}          & 0.404       & 0.386       & \green{0.555}             \\ 
\cline{2-13}
                                                          & \textbf{MU}    & \textbf{0.585}       & -           & 0.395          & \textbf{0.644}          & \textbf{0.629}          & \textbf{\blue{\underline{0.732}}}       & \textbf{0.673}          & 0.323          & \textbf{\green{0.701}}       & \textbf{0.533}       & \textbf{0.592}             \\ 
\hline \hline
\multirow{2}{*}{\textbf{\textit{Hi-En}}}                           & \textbf{MB}    & \textbf{0.444}       & \textbf{\blue{\underline{0.548}}}       & -              & 0.323          & 0.470          & \textbf{\green{0.501}}       & 0.420          & 0.360          & 0.395       & \textbf{0.464}       & \textbf{0.382}             \\ 
\cline{2-13}
                                                          & \textbf{MU}    & 0.425       & 0.494       & -              & \textbf{0.326}          & \textbf{\blue{0.528}}          & 0.481       & \textbf{\green{0.508}}          & \textbf{0.389}          & \textbf{0.477}       & 0.438       & 0.379             \\ 
\hline \hline
\multirow{2}{*}{\textbf{\textit{Ka-En}}}                           & \textbf{MB}    & \textbf{0.500}       & \textbf{\blue{0.613}}       & 0.503          & -              & 0.536          & 0.505       & \green{0.619}          & 0.366          & 0.467       & 0.439       & 0.586             \\ 
\cline{2-13}
                                                          & \textbf{MU}    & 0.455       & 0.609       & \textbf{0.550}          & -              & \textbf{\green{0.649}}          & \textbf{0.583}       & \textbf{\blue{\underline{0.702}}}          & \textbf{0.391}          & \textbf{0.537}       & \textbf{0.442}       & \textbf{0.628}             \\ 
\hline \hline
\multirow{2}{*}{\textbf{\textit{Ma-En}}}                           & \textbf{MB}    & \textbf{\green{0.572}}       & \textbf{0.545}       & 0.534          & 0.485          & -              & \textbf{0.552}       & \textbf{\blue{\underline{0.627}}}          & 0.292          & 0.493       & \textbf{0.466}       & 0.569             \\ 
\cline{2-13}
                                                          & \textbf{MU}    & 0.499       & 0.496       & \textbf{0.559}          & \textbf{0.500}          & -              & 0.511       & 0.604          & \textbf{0.340}          & \textbf{\green{0.563}}       & 0.464       & \textbf{\blue{0.603}}             \\ 
\hline \hline
\multirow{2}{*}{\textbf{\textit{Mr}}}                              & \textbf{MB}    & 0.406       & \blue{0.661}       & 0.400          & 0.580          & 0.430          & -           & \green{0.606}          & \textbf{0.523}          & 0.472       & 0.394       & 0.599             \\ 
\cline{2-13}
                                                          & \textbf{MU}    & \textbf{0.552}       & \textbf{\green{0.813}}       & \textbf{0.434}          & \textbf{0.681}          & 0.595          & -           & \textbf{0.703}          & 0.403          & \textbf{0.710}       & \textbf{0.468}       & \textbf{\blue{\underline{0.816}}}             \\ 
\hline \hline
\multirow{2}{*}{\textbf{\textit{Ta-En}}}                           & \textbf{MB}    & \textbf{0.511}       & \textbf{0.592}       & 0.509          & \green{0.593}          & \blue{0.614}          & 0.564       & -              & 0.340          & 0.450       & 0.430       & \blue{0.614}             \\ 
\cline{2-13}
                                                          & \textbf{MU}    & 0.459       & 0.516       & \textbf{0.567}          & \textbf{\blue{\underline{0.634}}}          & \textbf{0.621}          & \textbf{0.572}       & -              & \textbf{0.355}          & \textbf{0.513}       & \textbf{0.442}       & \textbf{\green{0.627}}             \\ 
\hline \hline
\multirow{2}{*}{\textbf{\textit{Ur-En}}}                           & \textbf{MB}    & \textbf{0.326}       & \textbf{0.382}       & 0.315          & 0.467          & 0.404          & \textbf{\green{0.468}}       & \textbf{\blue{\underline{0.484}}}          & -              & 0.298       & 0.296       & \textbf{0.359}             \\ 
\cline{2-13}
                                                          & \textbf{MU}    & 0.317       & 0.322       & \textbf{0.354}          & \textbf{\blue{0.473}}          & \textbf{\green{0.420}}          & 0.399       & 0.464          & -              & \textbf{0.347}       & \textbf{0.302}       & 0.340             \\ 
\hline \hline
\multirow{2}{*}{\textbf{\textit{Ur}}}                              & \textbf{MB}    & 0.377       & \green{0.655}       & \textbf{0.357}          & 0.597          & 0.426          & 0.548       & 0.652          & 0.404          & -           & 0.342       & \blue{0.678}             \\ 
\cline{2-13}
                                                          & \textbf{MU}    & \textbf{0.580}       & \textbf{\blue{\underline{0.804}}}       & 0.339          & \textbf{0.652}          & \textbf{0.635}          & \textbf{\green{0.803}}       & \textbf{0.669}          & \textbf{0.453}          & -           & \textbf{0.462}       & \textbf{0.753}             \\ 
\hline \hline
\multirow{2}{*}{\textbf{\textit{En}}}                              & \textbf{MB}    & 0.352       & \green{0.626}       & \textbf{0.481}          & 0.577          & 0.398          & 0.397       & 0.528          & 0.324          & 0.350       & -           & \blue{0.668}             \\ 
\cline{2-13}
                                                          & \textbf{MU}    & \textbf{0.606}       & \textbf{\blue{\underline{0.826}}}       & 0.476          & \textbf{0.656}          & \textbf{0.443}          & \textbf{\green{0.759}}       & \textbf{0.638}          & \textbf{0.354}          & \textbf{0.729}       & -           & \textbf{0.732}             \\
\hline
\end{tabular}
\caption{Macro F1 score of \textbf{MTx0}. Row-wise language names represent the target language, and column-wise language names represent the source language model. The better among \textbf{MB} and \textbf{MU} is highlighted in bold. \blue{\underline{Blue}} denotes the best source + model pair for a target language. \blue{Blue} represents the best, and \green{green} the second-best performing model per row (\textbf{MU} or \textbf{MB}).}
\label{tab:zeroshotModelTrans}
\end{table*}

\begin{table*}
\scriptsize
\centering
{\begin{tabular}{|c|c|c|c|c|c|c|c|c|c|c|c|c|c|}
\hline
\textbf{Lang}                   & \textbf{Model}               & \textbf{Shot} & \textbf{\textit{Bn}} & \textbf{\textit{Hi}} & \textbf{\textit{Hi-En}} & \textbf{\textit{Ka-En}} & \textbf{\textit{Ma-En}} & \textbf{\textit{Mr}} & \textbf{\textit{Ta-En}} & \textbf{\textit{Ur-En}} & \textbf{\textit{Ur}} & \textbf{\textit{En}} & \textbf{\textit{AllBOne}} \\ \hline
\multirow{4}{*}{\textbf{\textit{Bn}}}    & \multirow{2}{*}{\textbf{MB}} & \textbf{MTx32}           & -           & 0.705       & \textbf{0.647}          & 0.686          & 0.664          & \green{0.726}       & 0.705          & 0.579          & \blue{0.739}       & 0.71        & 0.725            \\ \cline{3-14} 
                                &                              & \textbf{MTx64}           & -           & (+0.055)    & (+0.033)       & (+0.034)       & (+0.046)       & (+0.024)    & (+0.045)       & (+0.071)       & (+0.011)    & (+0.04)     & (+0.015)         \\ \cline{2-14} 
                                & \multirow{2}{*}{\textbf{MU}} & \textbf{MTx32}           & -           & \textbf{\blue{\underline{0.777}}}       & 0.579          & \textbf{0.725}          & \textbf{0.712}          & \textbf{0.76}        & \textbf{0.718}          & \textbf{0.628}          & \textbf{\blue{\underline{0.777}}}       & \textbf{0.755}       & \textbf{\green{0.774}}            \\ \cline{3-14} 
                                &                              & \textbf{MTx64}           & -           & (+0.003)    & (+0.131)       & (+0.015)       & (+0.028)       & (+0.01)     & (+0.032)       & (+0.082)       & (+0.003)    & (+0.015)    & (-0.004)         \\ \hline \hline
\multirow{4}{*}{\textbf{\textit{Hi}}}    & \multirow{2}{*}{\textbf{MB}} & \textbf{MTx32}           & 0.656       & -           & \textbf{0.613}          & \blue{0.679}          & 0.465          & 0.672       & 0.647          & 0.569          & \green{0.677}       & 0.643       & 0.635            \\ \cline{3-14} 
                                &                              & \textbf{MTx64}           & (+0.014)    & -           & (+0.027)       & (+0.011)       & (+0.115)       & (+0.008)    & (+0.033)       & (+0.031)       & (+0.013)    & (+0.037)    & (+0.025)         \\ \cline{2-14} 
                                & \multirow{2}{*}{\textbf{MU}} & \textbf{MTx32}           & \textbf{0.739}       & -           & 0.607          & \textbf{0.696}          & \textbf{0.678}          & \textbf{0.732}       & \textbf{0.702}          & \textbf{0.624}          & \textbf{0.737}       &\textbf{\green{0.751}}       & \textbf{\blue{\underline{0.755}}}            \\ \cline{3-14} 
                                &                              & \textbf{MTx64}           & (+0.011)    & -           & (+0.063)       & (+0.014)       & (+0.032)       & (+0.028)    & (+0.018)       & (+0.056)       & (+0.013)    & (+0.009)    & (+0.015)         \\ \hline \hline
\multirow{4}{*}{\textbf{\textit{Hi-En}}} & \multirow{2}{*}{\textbf{MB}} & \textbf{MTx32}           & \textbf{0.538}       & \green{0.552}       & -              & 0.514          & 0.397          & \textbf{0.531}       & 0.538          & 0.487          & \textbf{0.544}       & \blue{0.567}       & 0.485            \\ \cline{3-14} 
                                &                              & \textbf{MTx64}           & (+0.012)    & (+0.018)    & -              & (+0.026)       & (+0.133)       & (+0.019)    & (+0.032)       & (+0.023)       & (+0.006)    & (+0.013)    & (+0.045)         \\ \cline{2-14} 
                                & \multirow{2}{*}{\textbf{MU}} & \textbf{MTx32}           & 0.521       & \textbf{0.553}       & -              & \textbf{\green{0.575}}          & \textbf{0.431}          & 0.53        & \textbf{0.552}          & \textbf{0.496}          & 0.532       & \textbf{\blue{\underline{0.581}}}       & \textbf{0.508}            \\ \cline{3-14} 
                                &                              & \textbf{MTx64}           & (+0.029)    & (+0.027)    & -              & (+0.055)       & (-0.051)       & (+0.02)     & (+0.038)       & (+0.054)       & (+0.028)    & (+0.019)    & (+0.042)         \\ \hline \hline
\multirow{4}{*}{\textbf{\textit{Ka-En}}} & \multirow{2}{*}{\textbf{MB}} & \textbf{MTx32}           & 0.589       & 0.604       & \textbf{0.585}          & -              & 0.606          & 0.57        & 0.598          & 0.53           & 0.584       & \green{0.61}        & \blue{0.624}            \\ \cline{3-14} 
                                &                              & \textbf{MTx64}           & (+0.031)    & (+0.036)    & (+0.045)       & -              & (+0.034)       & (+0.06)     & (+0.082)       & (+0.06)        & (+0.026)    & (+0.03)     & (+0.036)         \\ \cline{2-14} 
                                & \multirow{2}{*}{\textbf{MU}} & \textbf{MTx32}           & \textbf{0.618}       & \textbf{0.638}       & 0.551          & -              & \textbf{0.615}          & \textbf{0.572}       & \textbf{\green{0.664}}          & \textbf{0.561}          & \textbf{0.614}       & \textbf{0.62}        & \textbf{\blue{\underline{0.676}}}            \\ \cline{3-14} 
                                &                              & \textbf{MTx64}           & (+0.032)    & (+0.012)    & (+0.069)       & -              & (+0.005)       & (+0.058)    & (+0.036)       & (+0.029)       & (+0.036)    & (+0.02)     & (+0.004)         \\ \hline \hline
\multirow{4}{*}{\textbf{\textit{Ma-En}}} & \multirow{2}{*}{\textbf{MB}} & \textbf{MTx32}           & 0.597       & 0.562       & \textbf{0.566}          & 0.58           & -              & \textbf{0.58}        & \textbf{\blue{\underline{0.643}}}          & 0.484          & 0.567       & 0.582       & \textbf{\green{0.636}}            \\ \cline{3-14} 
                                &                              & \textbf{MTx64}           & (+0.033)    & (+0.068)    & (+0.054)       & (+0.05)        & -              & (+0.04)     & (+0.017)       & (+0.066)       & (+0.063)    & (+0.038)    & (+0.014)         \\ \cline{2-14} 
                                & \multirow{2}{*}{\textbf{MU}} & \textbf{MTx32}           & \textbf{0.6}         & \textbf{0.6}         & 0.537          & \textbf{0.618}          & -              & 0.555       & \green{0.624}          & \textbf{0.525}          & \textbf{\blue{0.627}}       & \textbf{0.589}       & 0.604            \\ \cline{3-14} 
                                &                              & \textbf{MTx64}           & (+0.04)     & (+0.04)     & (+0.083)       & (+0.032)       & -              & (+0.035)    & (+0.036)       & (+0.045)       & (+0.023)    & (+0.031)    & (+0.036)         \\ \hline \hline
\multirow{4}{*}{\textbf{\textit{Mr}}}    & \multirow{2}{*}{\textbf{MB}} & \textbf{MTx32}           & 0.73        & \blue{0.762}       & \textbf{0.629}          & 0.725          & 0.515          & -           & 0.74           & 0.617          & 0.729       & 0.708       & \green{0.757}            \\ \cline{3-14} 
                                &                              & \textbf{MTx64}           & (+0.03)     & (+0.038)    & (+0.041)       & (+0.035)       & (+0.225)       & -           & (+0.05)        & (+0.023)       & (+0.041)    & (+0.062)    & (+0.023)         \\ \cline{2-14} 
                                & \multirow{2}{*}{\textbf{MU}} & \textbf{MTx32}           & \textbf{0.784}       & \textbf{\green{0.834}}       & 0.608          & \textbf{0.753}          & \textbf{0.725}          & -           & \textbf{0.795}          & \textbf{0.701}          & \textbf{0.812}       & \textbf{0.801}       & \textbf{\blue{\underline{0.839}}}            \\ \cline{3-14} 
                                &                              & \textbf{MTx64}           & (+0.026)    & (+0.016)    & (+0.102)       & (+0.027)       & (+0.065)       & -           & (+0.035)       & (+0.039)       & (-0.002)    & (+0.019)    & (+0.001)         \\ \hline \hline
\multirow{4}{*}{\textbf{\textit{Ta-En}}} & \multirow{2}{*}{\textbf{MB}} & \textbf{MTx32}           & 0.608       & 0.619       & 0.587          & \green{0.631}          & \textbf{0.622}          & 0.607       & -              & 0.537          & 0.606       & 0.616       & \blue{0.661}            \\ \cline{3-14} 
                                &                              & \textbf{MTx64}           & (+0.012)    & (+0.001)    & (+0.013)       & (+0.009)       & (-0.002)       & (+0.003)    & -              & (+0.033)       & (+0.004)    & (+0.004)    & (-0.001)         \\ \cline{2-14} 
                                & \multirow{2}{*}{\textbf{MU}} & \textbf{MTx32}           & \textbf{0.628}       & \textbf{0.648}       & \textbf{0.592}          & \textbf{\green{0.654}}          & 0.61           & \textbf{0.621}       & -              & \textbf{0.547}          & \textbf{0.622}       & \textbf{0.619}       & \textbf{\blue{\underline{0.673}}}            \\ \cline{3-14} 
                                &                              & \textbf{MTx64}           & (+0.002)    & (+0.012)    & (+0.028)       & (+0.016)       & (+0.0)         & (+0.019)    & -              & (+0.043)       & (+0.018)    & (+0.011)    & (+0.007)         \\ \hline \hline
\multirow{4}{*}{\textbf{\textit{Ur-En}}} & \multirow{2}{*}{\textbf{MB}} & \textbf{MTx32}           & \textbf{0.518}       & 0.489       & 0.486          & \textbf{0.535}          & 0.461          & \textbf{\blue{\underline{0.559}}}       &\textbf{ 0.514}          & -              & \textbf{\green{0.55}}        & \textbf{0.516}       & \textbf{0.475}            \\ \cline{3-14} 
                                &                              & \textbf{MTx64}           & (+0.022)    & (+0.021)    & (+0.034)       & (+0.025)       & (+0.009)       & (+0.011)    & (+0.016)       & -              & (+0.01)     & (+0.014)    & (+0.005)         \\ \cline{2-14} 
                                & \multirow{2}{*}{\textbf{MU}} & \textbf{MTx32}           & 0.454       & \textbf{0.5}         & \textbf{0.491}          & 0.501          & \textbf{\blue{0.506}}          & 0.461       & \green{0.502}          & -              & 0.458       & 0.492       & 0.448            \\ \cline{3-14} 
                                &                              & \textbf{MTx64}           & (+0.016)    & (+0.03)     & (+0.009)       & (+0.009)       & (+0.024)       & (+0.059)    & (+0.018)       & -              & (+0.042)    & (+0.018)    & (+0.012)         \\ \hline \hline
\multirow{4}{*}{\textbf{\textit{Ur}}}    & \multirow{2}{*}{\textbf{MB}} & \textbf{MTx32}           & 0.785       & 0.782       & 0.669          & \textbf{0.75}           & 0.641          & \green{0.802}       & \textbf{\blue{0.803}}          & 0.623          & -           & 0.769       & 0.793            \\ \cline{3-14} 
                                &                              & \textbf{MTx64}           & (+0.055)    & (+0.038)    & (+0.061)       & (+0.05)        & (+0.109)       & (+0.028)    & (+0.027)       & (+0.057)       & -           & (+0.051)    & (+0.047)         \\ \cline{2-14} 
                                & \multirow{2}{*}{\textbf{MU}} & \textbf{MTx32}           & \textbf{\blue{\underline{0.839}}}       & \textbf{\green{0.83}}        & \textbf{0.686}          & 0.71           & \textbf{0.776}          & \textbf{0.808}       & 0.756          & \textbf{0.663}          & -           & \textbf{0.811}       & \textbf{0.824}            \\ \cline{3-14} 
                                &                              & \textbf{MTx64}           & (+0.011)    & (+0.01)     & (+0.024)       & (+0.05)        & (+0.034)       & (+0.012)    & (+0.054)       & (+0.067)       & -           & (+0.019)    & (+0.006)         \\ \hline \hline
\multirow{4}{*}{\textbf{\textit{En}}}    & \multirow{2}{*}{\textbf{MB}} & \textbf{MTx32}           & 0.768       & \blue{0.825}       & \textbf{0.763}          & \textbf{0.78}           & 0.695          & \green{0.811}       & \textbf{0.791}          & \textbf{0.686}          & 0.793       & -           & 0.798            \\ \cline{3-14} 
                                &                              & \textbf{MTx64}           & (+0.042)    & (+0.015)    & (+0.057)       & (+0.04)        & (+0.095)       & (+0.019)    & (+0.029)       & (+0.054)       & (+0.037)    & -           & (+0.022)         \\ \cline{2-14} 
                                & \multirow{2}{*}{\textbf{MU}} & \textbf{Mtx32}           & \textbf{0.815}       & \textbf{\blue{\underline{0.837}}}       & 0.656          & 0.754          & \textbf{0.727}          & \textbf{\green{0.829}}       & 0.79           & 0.633          & \textbf{0.824}       & -           & \textbf{0.801}            \\ \cline{3-14} 
                                &                              & \textbf{MTx64}           & (+0.015)    & (+0.013)    & (+0.074)       & (+0.026)       & (+0.073)       & (+0.011)    & (+0.02)        & (+0.117)       & (+0.016)    & -           & (+0.029)         \\ \hline
\end{tabular}}
 \caption{Macro F1 score of \textbf{MTx\{32, 64\}} for different source and target languages. Row-wise language names represent the target language, and column-wise language names represent the source language model. The better among \textbf{MB} and \textbf{MU} is highlighted in bold. \blue{\underline{Blue}} denotes the best source + model pair for a target language. \blue{Blue} represents the best, and \green{green} the second-best performing model per row (\textbf{MU} or \textbf{MB}). For \textbf{MTx64}, we show the gain added over \textbf{MTx32}.}
 \label{tab:fewshotModelTransfer}
\end{table*}

\subsubsection{Instance transfer}
Here we go the other way, i.e., instead of directly evaluating the model in the target language, we first translate the target language instances to the source language using Google Translate API\footnote{\url{https://cloud.google.com/translate}} on which the model is initially fine-tuned. In the zero shot setting \textbf{Ix0}, no gold target instances are used. In this scenario, we can again use a few gold target language instances (by translating to source language) to fine-tune the source model further. We name them \textbf{Ix32} and \textbf{Ix64}. As for the code-mixed instances, the translation across languages will be inaccurate; we limit this experiment to the monolingual setting only.

\subsubsection{Synthetic transfer}
Due to the less availability of low resource languages, in this technique, we experiment, can resource-rich language be useful if we translate them to low-resource language and build the model from scratch. To accomplish this, we translate (plentiful) gold source language (e.g., English) into the silver target language (e.g., Bengali, Hindi, etc.) instances using Google Translate API to train the model in the target language. This form of translation is widely used for model transfer; see Kozhevnikov and Titov~\cite{kozhevnikov2014cross}. Here, again, we can throw in zero or a few target gold instances, guiding to variations we name \textbf{STx0}, \textbf{STx32} and \textbf{STx64}.

\subsection{Experimental setup}
All models are assessed using the same 70:10:20 train, validation, and test split, stratified by class across the splits. For the transfer learning experiments, we use 32 and 64 training instances from each class to further fine-tune the model in other languages. We make three such different random sets for each target dataset to make our evaluation more effective and report the average performance.

All models were coded in Python, using the Pytorch library. The models were run for 10 epoch with Adam optimizer, batch\_size = 16, learning\_rate = $2e-5$ and adam\_epsilon = $1e-8$. We set the number of tokens $n$ = 128 for the experiments. We did not perform any hyper-parameter searching due to the limitation of computational resources; besides, the stated hyper-parameters have shown state-of-the-art performance in some of the previous shared-task~\cite{das2021abusive,banerjee2021exploring}. All our results are reported on the test set. For all the experiments, we use Ryzen 9, $5^\mathrm{th}$ gen 12 core processor, a Linux-based system with 64GB RAM and 16GB RTX 3080 GPU.

\subsection{Evaluation metric}
To remain consistent with existing literature, we assess our models in terms of \textbf{accuracy} and \textbf{macro F1-score}. These metrics together should be able to thoroughly evaluate the classification performance in distinguishing between abusive and normal posts. For zero-shot and few-shot settings, we report only \textbf{macro F1-score} due to scarcity of space.

\section{Results}
\subsection{Performance of ELFI}
In Table~\ref{tab:indRes}, we show the performance of each model for all the languages in terms of accuracy and macro F1 score. We observe model's performance varies from language to language. We see out of 10 language types, MuRIL (\textbf{MU}) outperforms m-BERT (\textbf{MB}) in 7 languages in terms of macro F1-score. Although \textbf{MU} was pre-trained in only 17 languages (quite less compared to \textbf{MB}), it was specifically focused on Indian languages. We posit that this is one of the primary reasons for its prevailing superior performance over \textbf{MB}.

\subsection{Performance of joint training/cross-lingual training}
Here we investigate the importance of joint training. Even though different language has different character symbols, the main intention for creating all these datasets have been to identify abusive post. Therefore, in order to check if the dataset in one language can be useful for another language or not, we execute joint training by merging all the training language instances. In Table \ref{tab:jointTraing} we summarize our results. We observe for some languages, the performance improved, and for some, the performance decreased. In general, \textbf{MU} works better than \textbf{MB} as observed in case of ELFI. Furthermore, we notice the performance deviation is negligible compared to the ELFI setting. Thus, as joint training is usually more expensive in terms of computational resource usage, it is more reasonable to perform self-training in a resource-constrained environment.

\subsection{Performance of model transfer}
\subsubsection{\textbf{Performance of MTx0}}
In Table \ref{tab:zeroshotModelTrans}, we show the performance of zero-shot model transfer outcomes across all pairs of languages. As expected, the macro F1 scores are worse compared to the ELFI setting. Further \textbf{MU} works better than \textbf{MB}. 

\begin{itemize}
    \item \textbf{\textit{Ur + MU}} is the most effective model when the target language is \textbf{\textit{Bn}}; \textbf{\textit{Hi + MU}} is the next most effective one. If \textbf{\textit{Hi}} is the target language, \textbf{\textit{Mr + MU}} is the most effective model followed by \textbf{\textit{Ur + MU}}. Likewise, when \textit{\textbf{\textit{Ur}}} is the target language, \textbf{\textit{Hi + MU}} model performs the best followed \textbf{\textit{Mr + MU}}. If \textbf{\textit{Mr}} is the target language although the \textbf{\textit{AllBOne + MU}} model performs the best, \textbf{\textit{Hi + MU}} is the closest second. This may be explained by their membership in the  \href{https://en.wikipedia.org/wiki/Indo-Aryan_languages}{Indo-Aryan language family}.
    \item The \textbf{\textit{Hi + MB}} model performs the best for the \textbf{\textit{Hi-En}} target language. This is possibly because of the fact that \textbf{\textit{Hi-En}} has the same underlying semantics as that of \textbf{\textit{Hi}} .
    \item \textbf{\textit{Ta-En + MU}} model is most effective for the \textbf{\textit{Ka-En}} language. Conversely \textbf{\textit{Ka-En + MU}} is the best source for the \textbf{\textit{Ta-En}} language. Besides, for the target language \textbf{\textit{Ma-En}}, \textbf{\textit{Ta-En + MB}} model obtains the highest performance. The effectiveness of these models is possibly because of the fact that all these languages belong to the \href{https://en.wikipedia.org/wiki/Dravidian_languages}{Dravidian language family}.
\end{itemize}

Overall we observe that zero-shot abusive language detection can be useful when the source and the target languages are from the same language family.

\subsubsection{\textbf{Performance of MTx32 and MTx64}}
Table \ref{tab:fewshotModelTransfer} illustrates the effect of allowing a small portion of available gold instances in the target language for second-stage fine-tuning. We observe adding more instances improves the overall performance in both \textbf{MU} and \textbf{MB} settings. With more target language gold instances, the deficit from ELFI continues to decrease. An additional interesting observation is that the combinations that were best in the zero-shot setting get altered in many cases here. 

\begin{itemize}
    \item Here, both \textbf{\textit{Ur + MU}} and \textbf{\textit{Hi + MU}} models perform best for target language \textbf{\textit{Bn}}. Conversely, \textbf{\textit{Bn + MU}} is the best source for the target language \textbf{\textit{Ur}}.
    \item For the target languages \textbf{\textit{Hi}}, \textbf{\textit{Mr}}, \textbf{\textit{Ka-En}}, and \textbf{\textit{Ta-En}}, we observe \textbf{\textit{AllBOne + MU}} model is the most effective. This could be due to the advantages drawn by this model from the two stage fine-tuning. While in the first stage it learns the diversity associated with the languages from different linguistic families, in the second stage it learns the target language specific intricacies from the few-shot labels provided.   
      \item Further we notice that for the target language \textbf{\textit{Hi-En}}, \textbf{\textit{En + MU}} model performs the best. This may be due to the \textbf{\textit{Hi-En}} examples shown to the \textbf{\textit{En + MU}} model  as few-shots from which it is able to learn the English-Hindi switching patterns and their semantic connections. 
    \item Likewise zero-shot performance, we observe that the \textbf{\textit{Ta-En + MB}} model achieves the best score on \textbf{\textit{Ma-En}}.
    \item The \textbf{\textit{Mr + MB}} model is most effective for the target language \textbf{\textit{Ur-En}}, followed by the \textbf{\textit{Ur}} model.
    \item For the target language \textbf{\textbf{\textit{En}}}, similar to zero-shot performance, \textbf{\textit{Hi + MU}} is the best source model.
\end{itemize}

Overall we observe that in a few shot setting, the \textbf{\textit{AllBOne}} model is the most effective. 

\begin{table}
\small
\begin{tabular}{|l|l|l|l|l|l|c|}
\hline
\textbf{Lang}                & \textbf{Model} & \textbf{\textit{Bn}} & \textbf{\textit{Hi}} & \textbf{\textit{Mr}} & \textbf{\textit{Ur}} & \textbf{\textit{En}} \\ \hline
\multirow{2}{*}{\textbf{\textit{Bn}}} & \textbf{MB}    & -           &\textbf{\blue{\underline{0.677}}}       & \textbf{0.59}        & \green{0.635}       &  \textbf{0.605}      \\ \cline{2-7} 
                             & \textbf{MU}    & -           & \green{0.621}       & 0.539       & \textbf{\blue{0.666}}       & 0.595       \\ \hline  \hline
\multirow{2}{*}{\textbf{\textit{Hi}}} & \textbf{MB}    & \textbf{\green{0.588}}       & -           & 0.519       & \green{0.588}       & \textbf{\blue{0.597}}        \\ \cline{2-7} 
                             & \textbf{MU}    & 0.53        & -           & \textbf{\blue{\underline{0.658}}}       & \textbf{\green{0.614}}       & 0.571       \\ \hline \hline
\multirow{2}{*}{\textbf{\textit{Mr}}} & \textbf{MB}    &\textbf{ 0.487}       & \textbf{\blue{\underline{0.617}}}       & -           & \green{0.53}        & 0.511        \\ \cline{2-7} 
                             & \textbf{MU}    & 0.445       & \blue{0.609}       & -           & \textbf{\green{0.568}}       & \textbf{0.519}       \\ \hline \hline
\multirow{2}{*}{\textbf{\textit{Ur}}} & \textbf{MB}    & \textbf{\green{0.563}}       & \textbf{\blue{\underline{0.627}}}       & 0.529       & -           & \textbf{0.524}        \\ \cline{2-7} 
                             & \textbf{MU}    & 0.498       & \blue{0.601}       & \textbf{\green{0.552}}       & -           & 0.507       \\ \hline
\end{tabular}
\caption{Macro F1 score of \textbf{Ix0}. Row-wise language names represent the target language, and column-wise language names represent the source language model. The better among \textbf{MB} and \textbf{MU} is highlighted in bold. \blue{\underline{Blue}} denotes the best source + model pair for a target language. \blue{Blue} represents the best, and \green{green} the second-best performing model per row (\textbf{MU} or \textbf{MB}).} 
\label{tab:zeroShotinstTransfer}
\end{table}

\begin{table*}
\small
\centering
\begin{tabular}{|c|c|c|c|c|c|c|c|}
\hline
\textbf{Lang}                & \textbf{Model}               & \textbf{Shot} & \textbf{\textit{Bn}} & \textbf{\textit{Hi}} & \textbf{\textit{Mr}} & \textbf{\textit{Ur}} & \textbf{\textit{En}} \\ \hline
\multirow{4}{*}{\textbf{\textit{Bn}}} & \multirow{2}{*}{\textbf{MB}} & \textbf{Ix32}   & -           & \green{0.679}       & 0.64        & \blue{0.695}       & \green{0.679}       \\ \cline{3-8} 
                             &                              & \textbf{Ix64}   & -           & (+0.011)    & (+0.02)     & (+0.015)    & (+0.031)    \\ \cline{2-8} 
                             & \multirow{2}{*}{\textbf{MU}} & \textbf{Ix32}   & -           & \textbf{\green{0.716}}       & \textbf{0.695}       & \textbf{\blue{\underline{0.724}}}       & \textbf{0.704}       \\ \cline{3-8} 
                             &                              & \textbf{Ix64}   & -           & (+0.004)    & (+0.015)    & (+0.006)    & (+0.016)    \\ \hline \hline
\multirow{4}{*}{\textbf{\textit{Hi}}} & \multirow{2}{*}{\textbf{MB}} & \textbf{Ix32}   & 0.661       & -           & \blue{0.691}       & 0.66        & \green{0.685}       \\ \cline{3-8} 
                             &                              & \textbf{Ix64}   & (+0.009)    & -           & (+0.009)    & (+0.01)     & (+0.015)    \\ \cline{2-8} 
                             & \multirow{2}{*}{\textbf{MU}} & \textbf{Ix32}   & \textbf{0.697}       & -           & \textbf{\blue{\underline{0.732}}}       & \textbf{0.717}       & \textbf{\green{0.719}}       \\ \cline{3-8} 
                             &                              & \textbf{Ix64}   & (+0.023)    & -           & (+0.008)    & (+0.013)    & (+0.001)    \\ \hline \hline
\multirow{4}{*}{\textbf{\textit{Mr}}} & \multirow{2}{*}{\textbf{MB}} & \textbf{Ix32}   & 0.624       & \green{0.678}       & -           & 0.653       & \blue{0.689}       \\ \cline{3-8} 
                             &                              & \textbf{Ix64}   & (+0.046)    & (+0.002)    & -           & (+0.027)    & (+0.021)    \\ \cline{2-8} 
                             & \multirow{2}{*}{\textbf{MU}} & \textbf{Ix32}   & \textbf{0.683}       & \textbf{\blue{\underline{0.742}}}       & -           & \textbf{\green{0.732}}       & \textbf{0.696}       \\ \cline{3-8} 
                             &                              & \textbf{Ix64}   & (+0.017)    & (-0.002)    & -           & (+0.008)    & (+0.014)    \\ \hline \hline
\multirow{4}{*}{\textbf{\textit{Ur}}} & \multirow{2}{*}{\textbf{MB}} & \textbf{Ix32}   & \blue{0.692}       & \green{0.685}       & 0.659       & -           & \textbf{0.652}       \\ \cline{3-8} 
                             &                              & \textbf{Ix64}   & (+0.038)    & (+0.025)    & (+0.021)    & -           & (+0.038)    \\ \cline{2-8} 
                             & \multirow{2}{*}{\textbf{MU}} & \textbf{Ix32}   & \textbf{\blue{\underline{0.722}}}       & \textbf{\green{0.706}}       & \textbf{0.674}       & -           & 0.644       \\ \cline{3-8} 
                             &                              & \textbf{Ix64}   & (+0.018)    & (+0.024)    & (+0.016)    & -           & (+0.046)    \\ \hline
\end{tabular}
 \caption{Macro F1 score of \textbf{Ix\{32, 64\}} for different source and target languages. Row-wise language names represent the target language, and column-wise language names represent the source language model. The better among \textbf{MB} and \textbf{MU} is highlighted in bold. \blue{\underline{Blue}} denotes the best source + model pair for a target language. \blue{Blue} represents the best, and \green{green} the second-best performing model per row (\textbf{MU} or \textbf{MB}). For \textbf{Ix64}, we show the gain over \textbf{Ix32}.}
\label{tab:fewshotInstTransfer}
\end{table*}

\subsection{Performance of instance transfer}
In this section, we compare the performance of instance transfer (\textbf{Ix}) with model transfer (\textbf{MTx}). Table \ref{tab:zeroShotinstTransfer} shows the performance of \textbf{Ix0}. We observe, unlike \textbf{MTx}, \textbf{MB} performs relatively better than \textbf{MU}.  Although comparing Table \ref{tab:zeroshotModelTrans} and \ref{tab:zeroShotinstTransfer}, we notice \textbf{MTx0} outperforms \textbf{Ix0} for all the languages. We observe two reasons for the inferior performance of \textbf{Ix0} -- (a) while translating one language to another, translation errors are inevitable and (b) translation tones down the level abusiveness of a post. 

Further, we also experiment with the few-shot setup in instance transfer environment (i.e., \textbf{Ix16} and \textbf{Ix32}). Table \ref{tab:fewshotInstTransfer} shows the performance, which although is better than \textbf{Ix0}, cannot outperform the numbers obtained from the model transfer schemes (see Table \ref{tab:fewshotModelTransfer} vs. Table \ref{tab:fewshotInstTransfer}). Overall we maintain that for abusive language detection, model transfer schemes are superior to instance transfer schemes.

\begin{table}
\small
\begin{tabular}{|c|c|c|c|c|}
\hline
\multicolumn{1}{|l|}{\textbf{Lang}} & \multicolumn{1}{l|}{\textbf{Model}} & \multicolumn{1}{l|}{\textbf{STx0}} & \multicolumn{1}{l|}{\textbf{STx32}} & \multicolumn{1}{l|}{\textbf{STx64}} \\ \hline
\multirow{2}{*}{\textbf{\textit{Bn}}}        & \textbf{MB}                         & 0.572                              & 0.726                               & 0.761                               \\ \cline{2-5} 
                                    & \textbf{MU}                         & \textbf{0.612}                              & \textbf{0.739}                               & \textbf{0.779}                               \\ \hline
\multirow{2}{*}{\textbf{\textit{Hi}}}        & \textbf{MB}                         & 0.601                              & \textbf{0.672}                               & \textbf{0.705}                               \\ \cline{2-5} 
                                    & \textbf{MU}                         & \textbf{0.617}                              & 0.666                               & 0.674                               \\ \hline
\multirow{2}{*}{\textbf{\textit{Mr}}}        & \textbf{MB}                         & \textbf{0.725}                              & 0.779                               & 0.800                               \\ \cline{2-5} 
                                    & \textbf{MU}                         & 0.698                              & \textbf{0.802}                               & \textbf{0.807}                               \\ \hline
\multirow{2}{*}{\textbf{\textit{Ur}}}        & \textbf{MB}                         & 0.654                              & 0.774                               & 0.821                               \\ \cline{2-5} 
                                    & \textbf{MU}                         & \textbf{0.664}                              & \textbf{0.810}                               & \textbf{0.833}                               \\ \hline
\end{tabular}
\caption{Macro F1 for \textbf{STx\{0,32,64\}}. Best performance is highlighted in bold.}
\label{tab:synthZeroFew}
\end{table}

\subsection{Performance of synthetic transfer}
Here we explore the significance of silver instances in a low-resource setting. In the scarcity of ample target instances for ELFI style training, we can leverage potentially plentiful instances from a source language, i.e., \textit{\textbf{\textit{En}}}. Table \ref{tab:synthZeroFew} shows the results. Even after using all available silver target instances, we observe that adding gold target instances gives steady improvements. This implies the worth of always having at least some gold target instances in such tasks.

\section{Error analysis}

To delve deeper, we conduct an error analysis on both models using a small number of test data points wrongly classified by the models. We examine our error from three independent directions -- \textbf{general error}, \textbf{instance transfer error}, and \textbf{synthetic transfer error}.

\if{0}
In order to understand the model performance we have considered stratified sampled small amount of data and their incorrect predicted labels by those two models. We have viewed this analysis from three independent directions such as from generic observation, instance transfer observation and synthetic transfer observation. From generic statements we were able to segregate those mainly into four categories.
\fi

\begin{figure*}
  \includegraphics[width=0.7\textwidth]{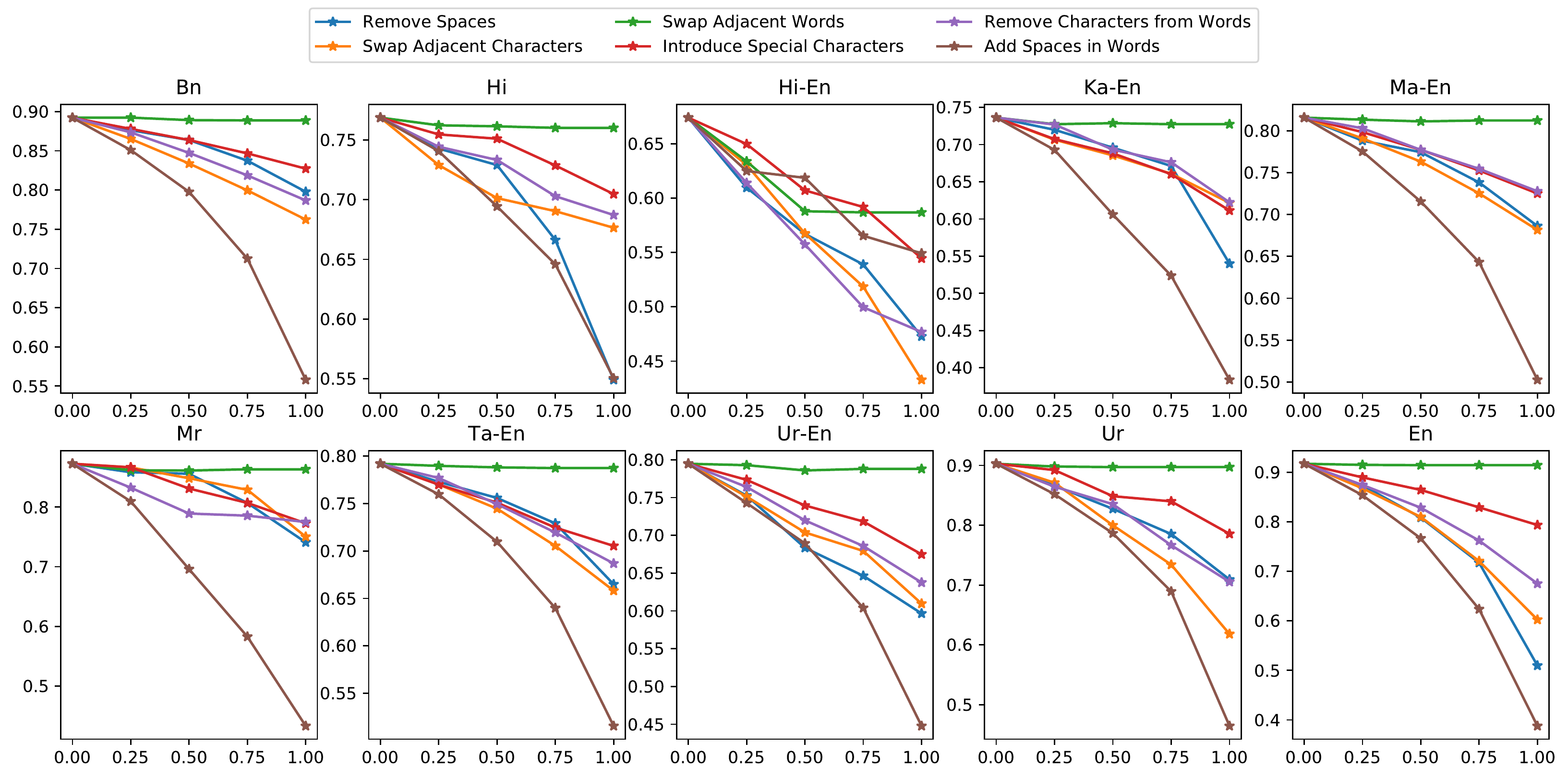}
  \caption{Macro F1 score under the adversarial attack. The X-axis represents (\%) of attack; Y-axis represents changes in macro F1 score.}
  \label{fig:advAttack}
\end{figure*}

\noindent \textbf{General error}: We analyze the common errors and segregate them into the following four categories.

\begin{itemize}
    \item The \textbf{presence of discriminatory features} is one of the crucial points of consideration for labeling a post abusive. We observe that models misclassify such posts where some specific words are present in the post that can be used in both abusive and non-abusive contexts. For example, the word `nigga' in a sentence is more likely to be considered abusive, whereas African-American people use the word in a non-abusive context. For instance, in the following tweet, ``Niggas pulled up on a nigga said hey come to HR for a drug test. Homie said word bet I quit LMFAOOOO" is predicted as abusive by the model, but its actual label is normal. On the other hand ``I play wit pussy not these niggas" is correctly predicted abusive by the model. We see m-BERT suffers mostly for this type of error.
    \item If the \textbf{contextual information} is limited in a post, models cannot predict its actual label. We observe some posts with no abusive text; however, its context makes it abusive possibly because of the presence of emoji, or attached URL. For example, the Bengali instance ``Ai maa*i tui holo akta \includegraphics[height=1em]{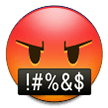} \includegraphics[height=1em]{Figures/angryFace.png} \includegraphics[height=1em]{Figures/angryFace.png} \includegraphics[height=1em]{Figures/angryFace.png}\footnote{We write non-English posts using Roman scripts.\label{footnoteGen}}" (Translation: You woman, you are a \includegraphics[height=1em]{Figures/angryFace.png} \includegraphics[height=1em]{Figures/angryFace.png} \includegraphics[height=1em]{Figures/angryFace.png} \includegraphics[height=1em]{Figures/angryFace.png}) without the emoji can be considered as non abusive; but the presence of the emojis makes it abusive. Our analysis identified that \textbf{MU} suffers mostly for this category of errors.
    \item \textbf{Implicit abusiveness} is another form of an error where the model fails to capture the context due to the absence of explicit abusive content. Understanding these instances requires understanding of sarcasm, complicated reasoning skills, or background knowledge of some situations. For instance the Hindi post ``Khujliwal ko karbwakar ashabaadi bana do \footnotemark[\getrefnumber{footnoteGen}]" (Translation: Make Khujliwal optimistic by getting it done.) is abusive toward a politician (Kejriwal) in India; nevertheless, the model fails to predict its actual label as it is sarcastic and need background knowledge about the politician. We notice \textbf{MB} suffers mostly for this category of errors.
    \item \textbf{Tentatively disputed annotations} are another type of situation where the ground truth label is ambiguous.  The context of a post can be multi-dimensional, and based on annotators' understanding, the labeling can be biased toward a specific direction. For example,  the following Bengali post ``Kon kon bokach*da mile man of the match nirbachan korche.\footnotemark[\getrefnumber{footnoteGen}]" (Translation: Which idiot-fucker is choosing Man of the Match.) has been wrongly annotated as normal, though the use of slur words makes it a clearly abusive post.
\end{itemize}

\noindent \textbf{Instance transfer error}: In the case of instance transfer, where we translate the target language model to the source language model, we observe the translation affects the true semantics of the actual target posts. Usually, the translator tool reduces the toxicity score of a sentence while translating; sometimes, it cannot translate appropriately because of context-sensitive situations. Hence model performance deteriorates while predicting those instances. For example, ``Tui akta kukurer bachha\footnotemark[\getrefnumber{footnoteGen}]" (a slur in Bengali) has been translated to ``You're a puppy" and does not look toxic.

\noindent \textbf{Synthetic transfer error}: A similar observation holds for synthetic transfer experiments. Here, the model is trained primarily on relatively less intense words. Hence, different kinds of abusive words/phrases get translated to either the same word or acquire some other representations. Naturally therefore the model performance degrades. For instance, when the following sentence ``Ladies and Gentlemen,  Victoria Soliz has fucked up once again" is translated to Hindi, it becomes ``Deviyon aur sajjanon, Viktoriya Solis ne ek baar phir gadbad kar di hai\footnotemark[\getrefnumber{footnoteGen}]" (Translation: Ladies and gentlemen, Victoria Solies has messed up once again). Though the original post was abusive, the translation made the sentence less intense and could be considered as normal.

\section{Robustness against adversarial changes in the text}
Observing the exceptional performance of ELFI style training, in this section, we try to understand, even if we can build an immaculate abusive speech classification model, how robust these models are against various adversarial changes. Specifically, we propose the following six black-box `attacks', which assumes that the attackers do not know the details of the detection algorithm, hence only creating the best possible guess to avoid exposure.

\noindent\textit{Remove spaces}: The spaces between adjacent words are removed. As the word-based language models primarily rely on tokenizing the text into a sequence of words, removing all spaces leads to a single <unk> token. Thus, finding the token boundaries would be difficult for the model.

\noindent\textit{Add spaces in words}: Conversely, we introduce spaces within individual words in a text, making the word unrecognizable based on characters treated as word separators. 

\noindent\textit{Remove characters from words}: Here we remove the characters from words to introduce typos. These words will still be readable by a human, but the tokens are changed from the classifier's perspective. 

\noindent\textit{Introduce special characters}: Here we introduce random special characters within words to introduce typos.

\noindent\textit{Swap adjacent characters}: We introduce another form of typo within words by swapping adjacent characters.

\noindent\textit{Swap adjacent words}: The transformer-based model tries to understand the underlying meaning of a sentence based on the way words are ordered. So, we swap adjacent words to change the relative ordering of words.

\noindent\textbf{Observations}:
In Figure \ref{fig:advAttack}, we demonstrate the performance of all the language models under various adversarial attacks for \textbf{MB}\footnote{Similar observations hold for \textbf{MU} as well.}. We see the performance of all the language models drops gradually with the increasing amount of adversarial attacks. Except for \textbf{\textit{Hi-En}}, all other models exhibit robustness against \textbf{adjacent word swap}, which indicates that the presence of abusive words is itself a strong signal for the model to judge whether a text is abusive or not irrespective of the relative order of the words. All the language models suffer when spaces are introduced within words, making it difficult to understand the actual word boundaries for the model. Overall similar observations are found for other types of adversarial attacks, which opens up another dimension of work that needs to be addressed to strengthen the models against such adversarial attacks.

\section{Conclusion}
In this paper, we tried to perform multilingual abusive language detection for Indic languages. We used transformer-based models to develop classifiers for abusive speech identification, using eight different languages from 14 publicly available resources. We perform several experiments for multiple languages under various settings - ELFI, zero-shot learning, few-shot learning, model transfer, instance transfer, cross-lingual learning, etc. Overall, we noticed that model transfer obtains better performance than instance transfer; further model transfer is advantageous when the source model language and target language belong to the same language family. Further in a few-shot setting the \textbf{\textit{AllBOne}} model performs the best as it gains from both the fine-tuning stages; while in the first stage it learns universal features, in the second stage it learns language specific features. We observed for low-resource languages, though synthetic silver instances are helpful to build classifiers for abusive language detection, further fine-tuning the model with gold target instances shows steady improvements.

One of the main drawbacks we observed was that the model's performance decreased against adversarial attacks. We plan to improve the existing models to make them agnostic of the adversarial attacks. Further, we plan to create datasets in other regional languages using the knowledge we obtained from our experiments. 

\bibliographystyle{ACM-Reference-Format}
\bibliography{sample-base}

\end{document}